\documentclass[letterpaper]{article} 
\usepackage{aaai2027}
\nocopyright
\usepackage[hyphens]{url}  
\usepackage{graphicx} 
\usepackage{natbib}  
\usepackage{caption} 
\usepackage{algorithm}
\usepackage{algorithmic}
\usepackage{amsmath}
\usepackage{amssymb}

\usepackage{newfloat}
\usepackage{listings}
\DeclareCaptionStyle{ruled}{labelfont=normalfont,labelsep=colon,strut=off} 
\floatstyle{ruled}
\newfloat{listing}{tb}{lst}{}
\floatname{listing}{Listing}

\usepackage{booktabs}

\usepackage{multirow}

\title{OccluRank: Controllable Occlusion-Aware Layout-to-Image Generation by Adding Just an Ordinal Rank}
\author{
    Wenyang Hong\textsuperscript{\rm 1}\equalcontrib,
    Yuan Wang\textsuperscript{\rm 2}\equalcontrib,
    Yanbin Hao\textsuperscript{\rm 1}\corresponding,
    Lanqing Xue\textsuperscript{\rm 3},
    Ke Wang\textsuperscript{\rm 4},\\
    Xiang Wang\textsuperscript{\rm 5},
    Kuien Liu\textsuperscript{\rm 1,\rm 6},
    Richang Hong\textsuperscript{\rm 1}
}
\affiliations{
    \textsuperscript{\rm 1}Hefei University of Technology\\
    \textsuperscript{\rm 2}School of Cyber Science and Technology, University of Science and Technology of China\\
    \textsuperscript{\rm 3}LCFC\\
    \textsuperscript{\rm 4}ByteDance Inc.\\
    \textsuperscript{\rm 5}School of Artificial Intelligence and Data Science, University of Science and Technology of China\\
    \textsuperscript{\rm 6}Institute of Software, Chinese Academy of Sciences\\
    haoyanbin@hfut.edu.cn
}

\begin{document}

\maketitle

\begin{abstract}
Layout-to-image generation enables explicit spatial control through bounding-box layouts, yet bounding boxes specify only instance locations and cannot represent their occlusion order. Existing methods may rely on additional geometric conditions, employ complex inference procedures, or aggregate independently constructed instance representations without explicitly modeling their occlusion-dependent interactions. We propose \textbf{OccluRank}, a simple and controllable occlusion-aware layout-to-image framework that augments each bounding box with only one ordinal rank. OccluRank encodes the user-specified occlusion order through lightweight rank-based conditioning and introduces an \textbf{Order-aware Instance Interaction (OII)} module to jointly update rank-conditioned instance representations before aggregation. This allows the specified order to guide information exchange among occluding instances without additional geometric inputs or specialized inference-time optimization. We further construct \textbf{OccluLayout}, a synthetic training dataset whose occlusion order and amodal annotations are derived directly from known scene geometry rather than estimated from partially occluded images using auxiliary prediction models. For comprehensive evaluation, we introduce \textbf{OccluLayout-Bench}, which uses multiple multimodal large language model evaluators to assess instance presence, spatial layout, attributes, and occlusion order, together with FID for overall image quality. Experiments show that OccluRank more reliably preserves target instances, follows specified layouts, and realizes desired occlusion relationships while maintaining comparable attribute consistency and overall image quality.
\end{abstract}

\begin{links}
    \link{Code}{https://github.com/Wenyang-hong/OccluRank}
\end{links}

\section{Introduction}
Recent advances in layout-to-image generation~\cite{wang2025msdiffusion,peng2025muse,zhang2025creatilayout,wu2025ifadapter,dahary2024boundedattention} have enabled precise control over object placement and instance-level semantics through bounding-box layouts, supporting applications such as advertising design, content creation, and scene editing~\cite{li2023gligen,zheng2023layoutdiffusion,xie2023boxdiff,zhao2023loco,xiang2025instanceassemble}. However, bounding-box layouts remain insufficient for scenes involving object occlusion, such as crowded scenes, human-object interactions, and layered compositions. This limitation arises because bounding boxes specify only the spatial extent of each instance, without indicating its occlusion order relative to others. Consequently, existing methods may produce incorrect occlusion relationships, merged objects, or attribute confusion~\cite{agrawal2026seethrough3d,liang2025vodiff}. We refer to the task of jointly controlling object layouts and their occlusion order as \emph{\textbf{occlusion-aware layout-to-image generation}}.

\begin{figure}[!t]
\centering
\includegraphics[width=0.48\textwidth]{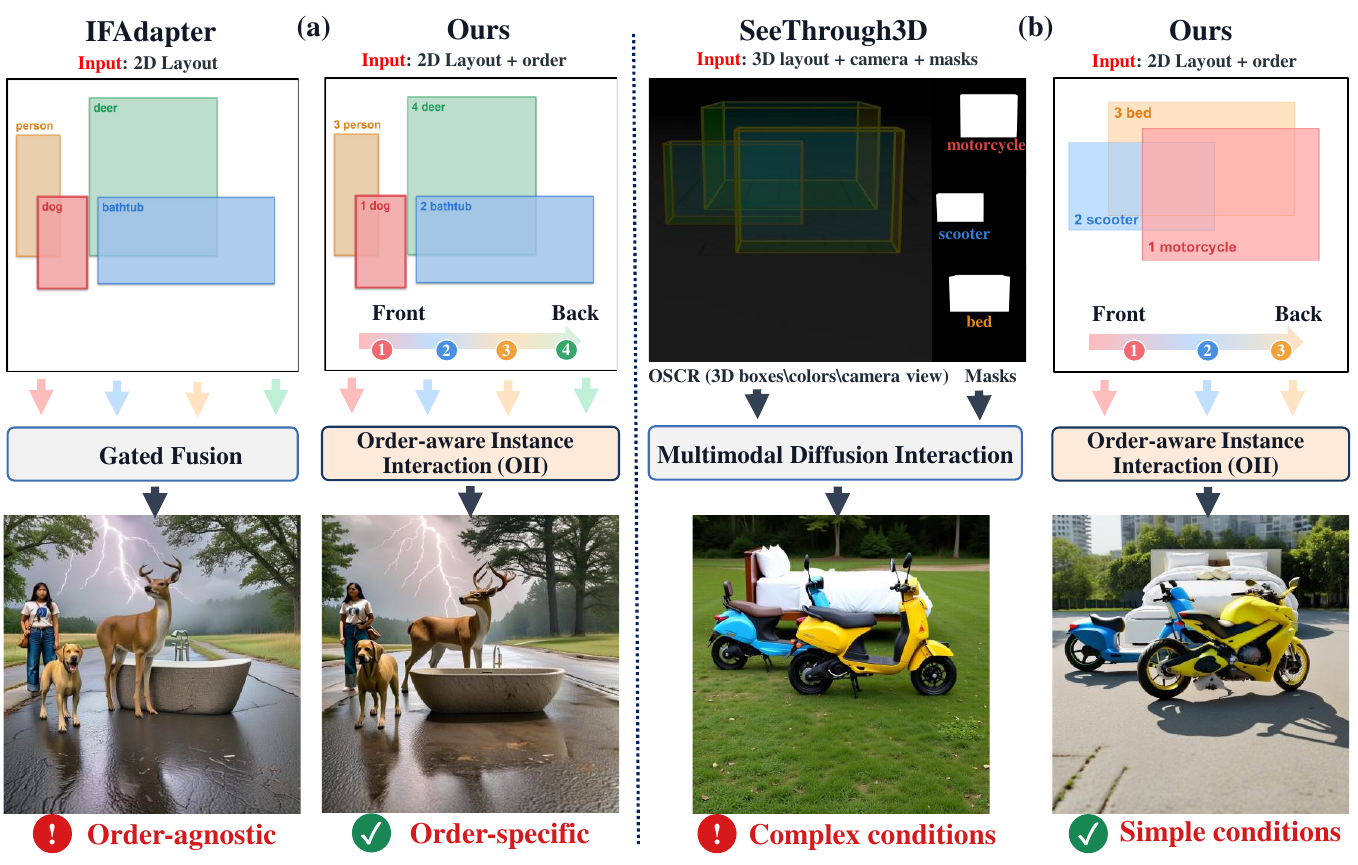}
\caption{Comparison with representative occlusion-control approaches. 
(a) IFAdapter is order-agnostic and cannot reliably produce the desired
occlusion order, whereas OccluRank uses ordinal ranks and OII for explicit
order control. 
(b) SeeThrough3D requires more complex geometric conditions, including an
Occlusion-Aware Scene Representation (OSCR) and instance masks, whereas
OccluRank assigns only one ordinal rank to each instance.
}
\label{fig:intro}
\end{figure}

This task poses two key challenges. \textbf{\textit{(1) Occlusion-dependent instance interaction:}} Existing methods~\cite{zhang2025creatilayout,xiang2025instanceassemble} improve object localization, instance--attribute binding, and semantic separation, but primarily isolate instance semantics rather than explicitly modeling their interactions under occlusion. IFAdapter~\cite{wu2025ifadapter} provides a natural basis for this task by independently constructing instance representations and combining them through gated semantic aggregation. However, this mechanism mainly regulates the contribution of each instance representation without explicitly modeling occlusion-dependent interactions, which may lead to ambiguous layering, missing instances, or cross-instance feature mixing, as shown in Figure~\ref{fig:intro}(a). \textbf{\textit{(2) Simple and controllable order specification:}} Effective interaction also requires explicit guidance on which instance should appear in front. However, IFAdapter is agnostic to user-specified occlusion order, leaving the resulting arrangement to the model's generative prior and thus limiting controllability. Recent methods introduce explicit order information through complex conditions or order-dependent inference mechanisms~\cite{agrawal2026seethrough3d,yang2025dccontrolnet,liang2025vodiff}. Methods based on complex conditions require additional inputs. SeeThrough3D, as exemplified in Figure~\ref{fig:intro}(b), further requires scene parameters and preprocessing, whereas inference-based methods rely on staged denoising and iterative latent optimization, resulting in a more involved and computationally demanding inference procedure. Therefore, incorporating user-specified occlusion order through a simple condition compatible with standard layout-to-image generation remains an open challenge.

To address these challenges, we introduce \textbf{OccluRank}, a controllable occlusion-aware layout-to-image framework that augments each bounding box with only one ordinal rank. OccluRank combines lightweight \textbf{rank-based order conditioning} with an \textbf{Order-aware Instance Interaction (OII)} module to support user-specified occlusion order and model interactions among occluding instances. Specifically, each instance is assigned an ordinal rank according to the user-specified occlusion order, in addition to the conventional global prompt, instance description, and bounding box. The rank is encoded as a learnable embedding and injected into the corresponding instance representation, enabling explicit order control without additional complex geometric inputs. OII then models occlusion-dependent interactions before feature aggregation. For each latent position covered by multiple instance bounding boxes, it organizes the corresponding rank-conditioned features as instance tokens and jointly updates them using an instance-wise Transformer. This allows the specified order to guide information exchange among occluding instances. As shown in Figure~\ref{fig:intro}, OccluRank better preserves the involved instances and more faithfully follows the specified occlusion order.


Beyond model design, effective training and evaluation require reliable occlusion supervision and detailed instance-level annotations. Existing public datasets either lack fine-grained instance
descriptions~\cite{agrawal2026seethrough3d} or rely on auxiliary models
to infer occlusion order and complete object geometry from partially
occluded images~\cite{li2026occlusionformer}.
Because the latter properties are not directly observable, the inferred
annotations may contain errors in occlusion relationships or hidden
object regions. We therefore construct \textbf{OccluLayout} from controllable 3D scenes whose geometry, object arrangement, and camera parameters are known before rendering. This enables occlusion order, amodal bounding boxes, and amodal masks to be derived directly from the complete scene geometry, together with fine-grained descriptions for each instance.

We further introduce \textbf{OccluLayout-Bench} for comprehensive evaluation. Unlike existing protocols that mainly assess occlusion order and may overlook missing, misplaced, or incorrectly rendered instances, OccluLayout-Bench evaluates five complementary aspects: instance presence with \textbf{Presence}, spatial layout adherence with \textbf{bounding-box mean Intersection over Union (Box mIoU)}, attribute consistency with \textbf{Color} and \textbf{Texture}, occlusion-order adherence with \textbf{Strict Pair} and \textbf{Strict Image}, and overall image quality with
\textbf{Fr\'echet Inception Distance
(FID)}~\cite{heusel2017gans}. Multiple multimodal large language models (MLLMs) independently evaluate instance presence, localization, attributes, and occlusion relations to assess robustness across evaluator choices, while FID is computed using the standard protocol. Experiments with three MLLM evaluators show that \textbf{OccluRank} more reliably preserves target instances, follows the specified layouts, and realizes the desired occlusion relationships under substantial occlusion, while maintaining comparable attribute consistency and overall image quality.

Our main contributions are summarized as follows:
\begin{itemize}
\item We propose \textbf{OccluRank}, a simple occlusion-aware layout-to-image framework that enables controllable occlusion order by adding only one ordinal rank to each bounding box. Its OII module further models occlusion-dependent interactions among instance representations. Experiments show that OccluRank better preserves target instances, follows specified layouts, and realizes desired occlusion orders while maintaining comparable attribute consistency and image quality.

\item We construct \textbf{OccluLayout}, a synthetic training dataset whose occlusion order and amodal annotations are derived directly from controllable scene geometry rather than estimated by auxiliary models. It also provides fine-grained descriptions for individual instances.

\item We introduce \textbf{OccluLayout-Bench}, which goes beyond order-only evaluation by jointly assessing instance presence, spatial layout, attributes, occlusion order, and image quality. Evaluations with multiple MLLMs further verify the robustness of the results across evaluator choices.
\end{itemize}

\section{Related Work}
\label{sec:related_work}

\paragraph{Layout-to-Image Generation.}
Layout-to-image generation introduces explicit spatial conditions into
diffusion models to control object placement and instance-level
semantics.
LayoutDiffusion~\cite{zheng2023layoutdiffusion} and
GLIGEN~\cite{li2023gligen} inject bounding-box layouts through
dedicated conditioning or attention layers.
MS-Diffusion~\cite{wang2025msdiffusion},
MUSE~\cite{peng2025muse}, and
CreatiLayout~\cite{zhang2025creatilayout} further incorporate
instance-level features or regional descriptions for fine-grained
multi-object control, while
InstanceAssemble~\cite{xiang2025instanceassemble} improves
layout-aware instance modeling through specialized attention.
Training-free approaches such as
BoxDiff~\cite{xie2023boxdiff},
LoCo~\cite{zhao2023loco}, and
Bounded Attention~\cite{dahary2024boundedattention} instead impose
bounding-box constraints by manipulating attention during inference. IFAdapter~\cite{wu2025ifadapter}, which provides the instance-conditioning pathway adopted in our framework, independently constructs spatially aligned maps for individual instances and aggregates overlapping maps using location-dependent scalar gates. While effective for spatial grounding, this mechanism only reweights independently constructed features and does not incorporate a user-specified visibility order or model order-conditioned interactions among overlapping instances.


\paragraph{Occlusion-Aware Layout-to-Image Generation.}
Recent occlusion-aware methods can be grouped into four categories according to how they incorporate visibility order. First, order-dependent inference methods modify the denoising process. For instance, VODiff~\cite{liang2025vodiff} combines sequential denoising with visibility-order-aware attention optimization, LayerBind~\cite{chen2026layer} constructs and fuses instance branches according to the desired layer order, and DepthArb~\cite{niu2026deptharb} resolves overlapping attention competition through depth-aware arbitration. These methods provide training-free control but require sequential branches or iterative attention and latent manipulation beyond standard sampling. Second, geometry-conditioned methods such as SeeThrough3D~\cite{agrawal2026seethrough3d} render translucent 3D bounding boxes from a specified viewpoint, providing explicit cues about hidden regions and camera geometry at the cost of richer 3D inputs. Third, methods based on multiple control inputs, such as DC-ControlNet~\cite{yang2025dccontrolnet}, combine point, box, or mask layouts with additional conditions and use spatial and layer reweighting to merge multiple objects according to a specified order. Fourth, rendering-based methods establish foreground priority by compositing decoupled instance representations. LaRender~\cite{zhan2025larender} uses transmittance-weighted latent rendering, while OcclusionFormer~\cite{li2026occlusionformer} learns density, opacity, and transmittance for Z-order-aware volumetric composition with additional amodal mask supervision.

\paragraph{Occlusion-Aware Dataset and Evaluation.}
Occlusion-aware generation requires reliable order and amodal supervision. COCOA~\cite{zhu2017semantic} and InstaOrder~\cite{lee2022instance} provide manually annotated amodal regions or pairwise occlusion and depth relations, but inherit COCO's low resolution and closed-set vocabulary and lack fine-grained instance descriptions. Recent high-resolution layout datasets provide open-vocabulary regional descriptions~\cite{zhang2025eligen,zhang2025creatilayout,li2025seg2any}, but do not jointly contain geometry-derived occlusion order and amodal annotations. SA-Z~\cite{li2026occlusionformer} obtains such annotations using auxiliary prediction models, whereas SeeThrough3D~\cite{agrawal2026seethrough3d} derives visibility order from controllable 3D scenes but provides only category-level descriptions. In contrast, our OccluLayout derives order and amodal annotations directly from complete scene geometry, while OccluLayout-Bench jointly evaluates instance presence, layout, attributes, occlusion order, and image quality.

\section{Method}
\label{sec:method}

\begin{figure*}[t]
\centering
\includegraphics[width=\textwidth]{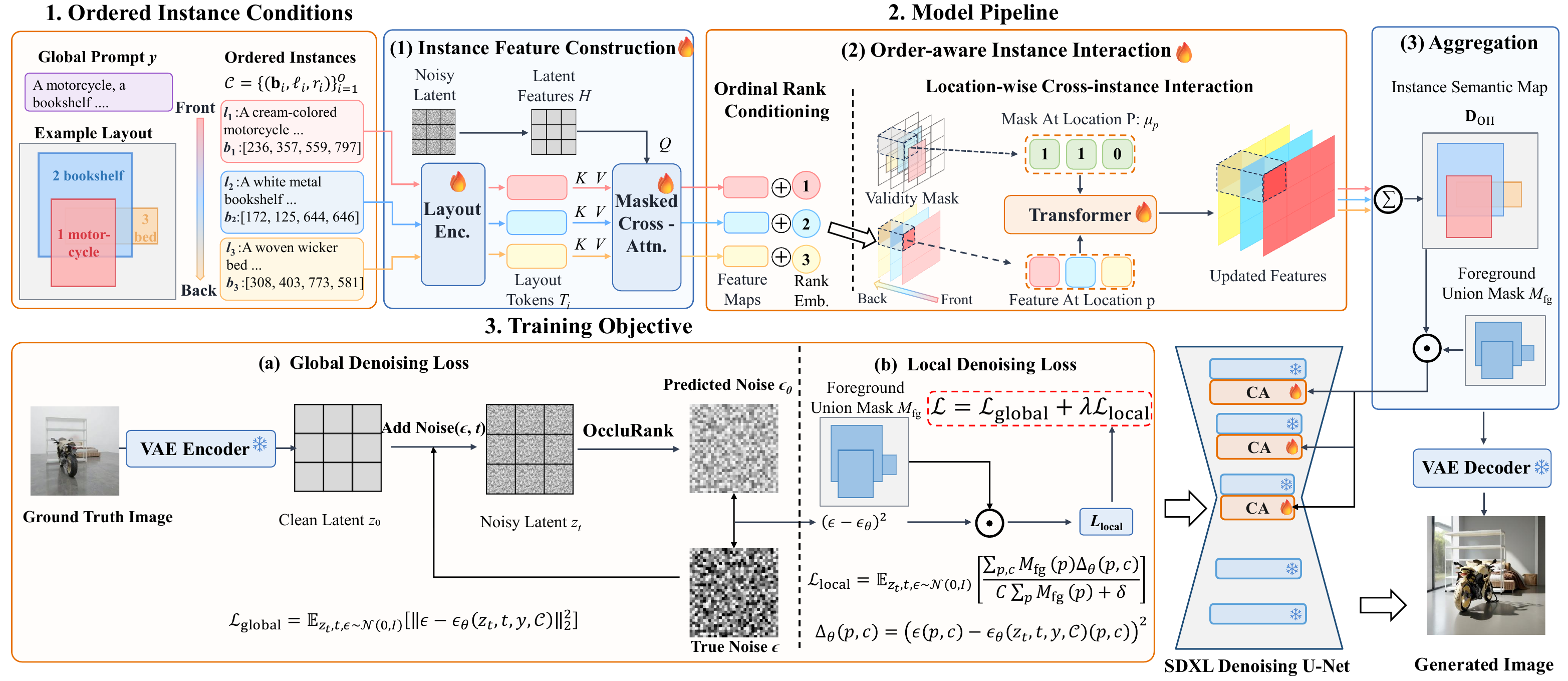}
\caption{
Overview of OccluRank. OccluRank constructs spatially aligned instance features, incorporates ordinal rank embeddings, and performs location-wise cross-instance interaction before aggregating into an Instance Semantic Map (ISM). The masked ISM is residually injected into selected cross-attention layers of SDXL.}
\label{fig:method}
\end{figure*}


\subsection{Overview and Problem Formulation}
\label{overview}

We build \textbf{OccluRank} upon a pretrained Stable Diffusion XL (SDXL) model~\cite{podell2024sdxl}, retaining its standard denoising pipeline while keeping the pretrained diffusion backbone frozen. OccluRank augments each instance condition with a single ordinal rank that specifies its intended position in a user-defined front-to-back occlusion order.


Given a global prompt $y$, the ordered layout condition is defined as
\begin{equation}
\mathcal{C}
=
\left\{
\left(\mathbf{b}_i,\ell_i,r_i\right)
\right\}_{i=1}^{O},
\qquad
\left\{r_i\right\}_{i=1}^{O}
=
\left\{1,\ldots,O\right\},
\label{eq:ordered_layout}
\end{equation}
where $O$ denotes the number of instances, $\ell_i$ is the description of instance $i$, and $\mathbf{b}_i=(x_i^0,y_i^0,x_i^1,y_i^1)$ is its 2D amodal bounding box. The ordinal rank $r_i$ specifies the position of instance $i$ in the front-to-back occlusion order, with a smaller value indicating a more foreground position. For any pair of overlapping instances, $r_i<r_j$ indicates that instance $i$ should appear in front of and occlude instance $j$. The ranks encode relative visibility order rather than metric depth. The objective is to generate an image that follows the global prompt, places each instance within its specified box, preserves its described attributes, and satisfies the rank-induced occlusion relations.


As illustrated in Figure~\ref{fig:method}, OccluRank follows an \emph{interaction-before-aggregation} pipeline. The layout encoder and masked instance cross-attention adopted from IFAdapter~\cite{wu2025ifadapter} first construct a spatially aligned feature map for each instance independently. Rather than directly aggregating these maps, OccluRank keeps them separated and injects the corresponding rank embeddings. At each latent location, the rank-conditioned features of all covering instances are organized along the instance dimension and jointly updated by the \textbf{Order-aware Instance Interaction} (\textbf{OII}) module. The updated features are subsequently aggregated into an Instance Semantic Map (ISM) and residually injected into selected cross-attention layers of SDXL.



\subsection{Per-Instance Feature Construction}
\label{sec:instance_grounding}

Before modeling occlusion relationships, we first construct an instance-specific feature representation that preserves both the semantic content and spatial support of each condition. Following IFAdapter~\cite{wu2025ifadapter}, this construction is performed independently for different instances, preventing their semantic information from being mixed before the subsequent order-aware interaction. These conditions are integrated into a compact sequence of layout tokens:
\begin{equation}
\mathbf{T}_i
=
\Phi_{\mathrm{layout}}
\left(
\mathbf{b}_i,\ell_i
\right)
\in\mathbb{R}^{L_T\times C},
\label{eq:layout_tokens}
\end{equation}
where $L_T$ and $C$ denote the number and feature dimension of the layout tokens, respectively.

Having obtained the layout tokens, we next integrate each instance condition into the current image representation through an instance-specific cross-attention pathway. Let
$\mathbf{H}\in\mathbb{R}^{N\times C}$ denote the current image features, where $N$ is the number of latent spatial locations. For each instance $i$, the image features provide the queries, while its layout tokens $\mathbf{T}_i$ provide the keys and values. The corresponding spatially aligned feature map is computed by the Masked-Attention~\cite{cheng2022masked}, that is
\begin{equation}
\begin{aligned}
\mathbf{F}_i=\operatorname{MaskAttn}\left(\mathbf{H}\mathbf{W}_{Q},\mathbf{T}_i\mathbf{W}_{K},\mathbf{T}_i\mathbf{W}_{V};\mathbf{M}_i\right).
\end{aligned}
\label{eq:instance_grounding}
\end{equation}
Here, $\mathbf{M}_i$ is an additive spatial attention mask derived from the bounding box of instance $i$. Let $\Omega(\mathbf{b}_i)$ denote the set of latent locations covered by $\mathbf{b}_i$. $\mathbf{M}_i(p,q\in\{1,\cdots,L_T\})= 0$ if $p\in\Omega(\mathbf{b}_i)$, and $ -\infty$ otherwise.

\subsection{Order-aware Instance Interaction}
\label{sec:oii}

The feature maps $\{\mathbf{F}_i\}_{i=1}^{O}$ preserve the semantics and spatial support of individual instances. However, they remain independent even when multiple bounding boxes cover the same latent location. Directly aggregating these features would collapse the instance dimension before their occlusion-dependent relationships are modeled, making it difficult to resolve semantic competition and visibility order afterward. OII addresses this aggregation bottleneck by first introducing the prescribed ordinal ranks, then allowing overlapping instance features to interact while they remain separated, and finally aggregating the updated representations.


\paragraph{Ordinal rank conditioning.} 
We maintain a learnable rank-embedding table
$\mathbf{E}_{\mathrm{rank}}\in\mathbb{R}^{O_{\max}\times C}$, where $O_{\max}$ is the maximum supported number of instances. The ordinal rank $r_i$ is encoded as
$\mathbf{e}_i^{r}=\mathbf{E}_{\mathrm{rank}}[r_i]$, and the rank-conditioned feature is defined as
\begin{equation}
\mathbf{X}_i(p)
=
\mu_i(p)
\left(
\mathbf{F}_i(p)
+
\mathbf{e}_i^{r}
\right),
\label{eq:rank_condition}
\end{equation}
where $\mu_i(p)=1$ indicates that instance $i$ covers latent location $p$. The rank embedding is broadcast over all valid locations of the instance, providing a consistent front-to-back role while setting features outside its bounding box to zero.



\paragraph{Location-wise cross-instance interaction.}
The central operation of OII is performed along the instance dimension rather than the spatial dimension. At each latent location $p$, we organize all instance slots into a sequence:
\begin{align}
\mathbf{X}_p
=
\left[
\mathbf{X}_1(p);
\ldots;
\mathbf{X}_O(p)
\right].
\label{eq:instance_sequence}
\end{align}
The rows of $\mathbf{X}_p$ represent different instances at the same spatial location, rather than different spatial tokens.


OII applies a Transformer block~\cite{vaswani2017attention} along the instance dimension:
\begin{equation}
    \widetilde{\mathbf{X}}_p=\operatorname{TransformerBlock}\left(\mathbf{X}_p;\boldsymbol{\mu}_p\right).
    \label{eq:oii}
\end{equation}
Here, $\boldsymbol{\mu}_p=\left[\mu_1(p),\ldots,\mu_O(p)\right]$ serves as the attention validity mask, and locations not covered by any instance are skipped. Through self-attention, each valid instance token receives a vector-valued update conditioned on the semantic content and ordinal roles of the other valid instances at the same location. Consequently, an instance feature can adapt its representation according to which objects compete with it and which of them should appear in front.

Unlike reweighting-based aggregation, OII directly updates instance representations before fusion. IFAdapter uses scalar gates to adjust independently constructed features, whereas DC-ControlNet~\cite{yang2025dccontrolnet} employs ordered cross-element Transformers to predict spatial and layer weights for reweighting the original element features. OII instead retains $\widetilde{\mathbf{X}}_p$ as the updated representation for subsequent aggregation, allowing each feature channel to adapt to the contents and ordinal roles of the other overlapping instances. Thus, the prescribed order affects the high-dimensional content of each instance representation rather than only its contribution to the aggregated feature.

\paragraph{Interaction-aware aggregation and injection.}
After interaction, the updated tokens are restored to their instance and spatial positions and aggregated into an Instance Semantic Map (ISM):
\begin{equation}
\mathbf{D}_{\mathrm{OII}}(p)
=
\sum_{i=1}^{O}
\mu_i(p)
\widetilde{\mathbf{X}}_i(p).
\label{eq:ism}
\end{equation}
Although the ISM is formed by summation, each contributing feature has already incorporated the ranks and contents of the other covering instances. The aggregation therefore preserves order-aware interactions without predicting density, opacity, or rendering weights.

To restrict the ISM to the layout-conditioned regions, we inject the resulting signal by a gated residual connection:
\begin{equation}
\mathbf{H}'
=
\mathbf{H}
+
\gamma
\mathbf{M}_{\mathrm{fg}}
\odot
\mathbf{D}_{\mathrm{OII}},
\qquad
\gamma=s\tanh(\alpha),
\label{eq:ism_injection}
\end{equation}
where $\mathbf{M}_{\mathrm{fg}}(p)=\mathbb{I}\left[\sum_{i=1}^{O}\mu_i(p)>0\right]\in\{0,1\}^{N\times 1}$ denotes the union mask of all instance regions, $\odot$ denotes element-wise multiplication, $s$ is the adapter scale, and $\alpha$ is learnable.




\subsection{Training Objective and Inference}
\label{sec:learning}

We initialize OccluRank from a pretrained SDXL model. The trainable components include the layout-conditioning pathway, ordinal rank embeddings, OII blocks, and residual injection parameters. Following the standard diffusion training objective~\cite{ho2020denoising}, the global denoising loss is defined as
\begin{equation}
\mathcal{L}_{\mathrm{global}}
=
\mathbb{E}_{z_t,t,\epsilon\sim\mathcal{N}(0,I)}
\left[
\left\|
\epsilon
-
\epsilon_{\theta}
\left(
z_t,t,y,\mathcal{C}
\right)
\right\|_2^2
\right],
\label{eq:global_loss}
\end{equation}
where $\epsilon_{\theta}$ denotes the noise-prediction network with trainable OccluRank parameters $\theta$, $z_t$ is the noisy latent at diffusion timestep $t$, and $\epsilon$ is the Gaussian noise used to construct $z_t$. 



The global objective distributes denoising supervision over the entire latent and may underemphasize the small regions occupied by the requested instances, including the overlap boundaries where their visibility order is expressed. We therefore introduce a region-focused denoising objective:
\begin{equation}
\mathcal{L}_{\mathrm{local}}
=
\mathbb{E}_{z_t,t,\epsilon\sim\mathcal{N}(0,I)}
\left[
\frac{
\sum_{p,c}
\mathbf{M}_{\mathrm{fg}}(p)
\Delta_{\theta}(p,c)
}{
C\sum_p
\mathbf{M}_{\mathrm{fg}}(p)
+
\delta
}
\right],
\label{eq:local_loss}
\end{equation}
where $\Delta_{\theta}(p,c)=\left(\epsilon(p,c)-\epsilon_{\theta}(z_t,t,y,\mathcal{C})(p,c)\right)^2$ ($c\in\{1, \cdots, C\}$) is the element-wise denoising error and $\delta$ is a small constant for numerical stability. By normalizing the denoising error over the box-union region, this objective provides stronger supervision for instance structures and occlusion boundaries. 

The complete objective is
\begin{equation}
\mathcal{L}=
\mathcal{L}_{\mathrm{global}}
+
\lambda\mathcal{L}_{\mathrm{local}},
\label{eq:total_loss}
\end{equation}
where $\lambda$ balances global image generation and region-focused denoising.

At inference time, the user provides a global prompt together with an ordered list of instance descriptions and 2D amodal bounding boxes. At each denoising step, OccluRank constructs the per-instance feature maps, performs rank-conditioned interaction through OII, and injects the resulting ISM into the selected SDXL layers. The remaining generation process follows standard SDXL sampling. Thus, OccluRank enables explicit occlusion control using only one ordinal rank per instance, without externally supplied masks, depth maps, camera parameters, or 3D conditions.

\section{Dataset and Benchmark Construction}
\label{data}

\begin{figure}[t]
    \centering
    \includegraphics[width=\columnwidth]{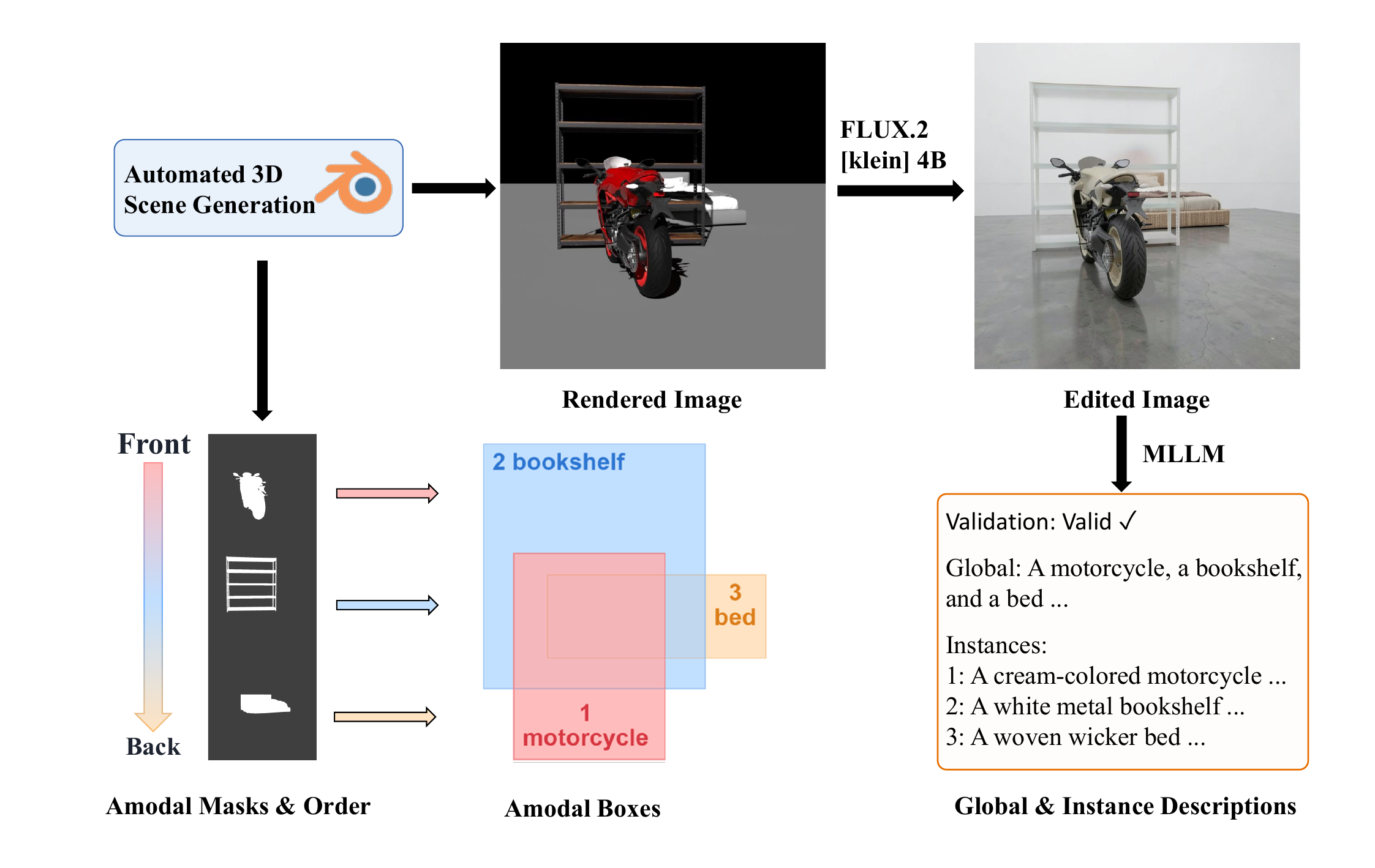}
    \caption{Overview of the OccluLayout construction.}
    \label{fig:dataset}
\end{figure}


\subsection{OccluLayout Dataset}
\label{subsec:occlulayout}

Reliable occlusion-aware generation requires training data that jointly provide instance-level descriptions, amodal spatial annotations, and explicit front-to-back order. Existing datasets do not fully satisfy these requirements. COCO-based resources such as COCOA~\cite{zhu2017semantic} and InstaOrder~\cite{lee2022instance} are limited by low-resolution images, class-level labels, and manual annotations. Recent high-resolution datasets provide open-vocabulary regional descriptions but generally lack explicit occlusion order or amodal geometry. SA-Z~\cite{li2026occlusionformer} obtains these annotations through auxiliary prediction, whereas SeeThrough3D~\cite{agrawal2026seethrough3d} derives order from geometry but provides only class-level descriptions and no instance-level amodal annotations. In contrast, OccluLayout combines high-resolution images and fine-grained instance descriptions with geometry-derived occlusion order, amodal boxes, and amodal masks.

\begin{table*}[t]
\centering
\small
\setlength{\tabcolsep}{1mm}

\begin{tabular}{@{}c|l|c|ccccccc@{}}
\toprule
Evaluator (MLLM)
& \multicolumn{1}{c|}{Method}
& \multicolumn{1}{c|}{Venue}
& Presence $\uparrow$
& Box mIoU $\uparrow$
& Color $\uparrow$
& Texture $\uparrow$
& Strict Pair $\uparrow$
& Strict Image $\uparrow$
& FID $\downarrow$ \\
\midrule

\multirow[c]{7}{*}{\shortstack[c]{Qwen3-VL-235B-\\A22B}}
& CreatiLayout
& ICCV'2025
& 87.92
& 53.83
& 81.06
& 71.70
& \underline{67.08}
& 50.78
& 62.754 \\

& LaRender
& ICCV'2025
& 55.14
& 16.88
& 46.25
& 38.41
& 16.63
& 7.90
& 96.556 \\

& OcclusionFormer$^{\dagger}$
& ICML'2026
& 76.49
& 50.36
& 70.45
& 57.89
& 55.28
& 39.77
& 72.707 \\

& OcclusionFormer
& ICML'2026
& 78.88
& 49.29
& 74.28
& 66.11
& 51.82
& 35.49
& \textbf{57.124} \\

& VODiff
& CVPR'2025
& 56.29
& 1.93
& 25.45
& 15.69
& 25.10
& 12.84
& 85.396 \\

& IFAdapter
& ICCV'2025
& \underline{90.17}
& \underline{60.80}
& \underline{83.46}
& \underline{74.84}
& 66.06
& \underline{51.55}
& 62.923 \\

& \textbf{Ours}
& --
& \textbf{91.58}
& \textbf{63.20}
& \textbf{85.23}
& \textbf{76.02}
& \textbf{76.45}
& \textbf{62.44}
& \underline{62.746} \\

\midrule

\multirow[c]{7}{*}{Qwen3-VL-32B}
& CreatiLayout
& ICCV'2025
& 88.07
& 54.93
& 78.59
& 71.51
& \underline{68.22}
& \underline{52.46}
& 62.754 \\

& LaRender
& ICCV'2025
& 55.29
& 17.32
& 45.30
& 42.82
& 18.50
& 9.59
& 96.556 \\

& OcclusionFormer$^{\dagger}$
& ICML'2026
& 77.05
& 50.43
& 70.48
& 61.34
& 54.60
& 37.18
& 72.707 \\

& OcclusionFormer
& ICML'2026
& 79.71
& 49.67
& 73.93
& 67.70
& 52.27
& 35.88
& \textbf{57.124} \\

& VODiff
& CVPR'2025
& 57.86
& 0.09
& 25.48
& 21.57
& 24.99
& 12.84
& 85.396 \\

& IFAdapter
& ICCV'2025
& \underline{91.44}
& \underline{60.87}
& \underline{82.59}
& \textbf{73.19}
& 66.46
& 51.42
& 62.923 \\

& \textbf{Ours}
& --
& \textbf{92.50}
& \textbf{64.23}
& \textbf{83.92}
& \underline{72.92}
& \textbf{77.07}
& \textbf{62.82}
& \underline{62.746} \\

\midrule









\multirow[c]{7}{*}{GLM-4.6V}
& CreatiLayout
& ICCV'2025
& 89.60
& 54.21
& 82.95
& 71.36
& \underline{62.66}
& \underline{47.41}
& 62.754 \\

& LaRender
& ICCV'2025
& 57.53
& 17.64
& 45.81
& 38.17
& 17.03
& 9.33
& 96.556 \\

& OcclusionFormer$^{\dagger}$
& ICML'2026
& 79.21
& 51.77
& 73.65
& 60.05
& 50.11
& 34.72
& 72.707 \\

& OcclusionFormer
& ICML'2026
& 81.04
& 49.89
& 76.18
& 65.54
& 46.03
& 30.31
& \textbf{57.124} \\

& VODiff
& CVPR'2025
& 59.10
& 21.22
& 28.57
& 18.32
& 20.55
& 9.86
& 85.396 \\

& IFAdapter
& ICCV'2025
& \underline{91.79}
& \underline{61.68}
& \underline{86.64}
& \underline{72.13}
& 59.93
& 44.69
& 62.923 \\

& \textbf{Ours}
& --
& \textbf{92.85}
& \textbf{64.13}
& \textbf{87.79}
& \textbf{73.23}
& \textbf{67.31}
& \textbf{52.46}
& \underline{62.746} \\

\bottomrule
\end{tabular}

\caption{
Quantitative comparison on OccluLayout-Bench using three MLLM evaluators.
VODiff is evaluated at $512{\times}512$ and all other methods at
$1024{\times}1024$.
$^{\dagger}$ denotes the official checkpoint without further training on
OccluLayout.
}
\label{tab:main_results}
\end{table*}

To construct \textbf{OccluLayout}, we generate controllable 3D scenes in Blender~\cite{blender2018} using public assets from 71 common object categories. As illustrated in Fig.~\ref{fig:dataset}, each scene contains two to five objects whose positions, scales, orientations, and camera viewpoints are varied to create diverse occlusion patterns, while physically intersecting scenes are rejected. For each accepted scene, we render a $1024\times1024$ RGB image and obtain each instance's amodal mask by rendering it separately with all other objects hidden. Its amodal bounding box is computed from the complete projected mask, and the front-to-back order is derived from the relative object positions in camera coordinates. 


To improve visual and semantic diversity, we use FLUX.2 [klein] 4B to edit scene backgrounds and selected object appearances while preserving their geometry and spatial arrangement. Qwen3-VL-235B-A22B-Instruct~\cite{bai2025qwen3vl} validates the edited images and generates global captions and fine-grained instance descriptions. Inter-object and visibility relations are excluded from instance descriptions to prevent leakage of the target order. The resulting collection contains 34,496 images and 117,433 instances.
Of these, 33,496 images form the OccluLayout training set, with 78.8\%
containing overlapping boxes. Each sample provides an RGB image, a global caption, instance descriptions, 2D amodal boxes, a front-to-back ordered instance list, and auxiliary amodal masks. 


\subsection{OccluLayout-Bench}
\label{subsec:bench}

Existing evaluations of occlusion-aware generation primarily assess pairwise visibility order. However, correct ordering alone does not ensure successful generation, as target instances may be missing, misplaced, or inconsistent with their specified attributes. We therefore introduce \textbf{OccluLayout-Bench}, comprising a randomly
held-out set of 1,000 images with 3,386 instances and an evaluation protocol
that jointly assesses instance presence, spatial adherence, attribute
consistency, occlusion order, and overall image quality.

Specifically, \textbf{Presence} measures whether each requested instance remains recognizable, while \textbf{Box mIoU} evaluates its alignment with the input box. \textbf{Color} and \textbf{Texture} assess instance-level attribute consistency, assigning zero credit to absent or unrecognizable instances. We further introduce two strict occlusion metrics. \textbf{Strict Pair} counts an overlapping pair as correct only when both instances are recognizable and their predicted front-to-back relation matches the specified order. \textbf{Strict Image} requires all overlapping pairs in an image to satisfy this criterion. FID~\cite{heusel2017gans} complements these structured metrics by measuring overall image quality. All semantic, spatial, attribute, and order metrics are independently evaluated using multiple MLLMs to reduce dependence on a single evaluator.

\section{Experiments}
\label{sec:experiments}

We train OccluRank on OccluLayout and evaluate it on OccluLayout-Bench. Experimental settings, implementation details and more results are provided in the Appendix.

\begin{figure*}[!t]
\centering
\includegraphics[width=0.96\textwidth]{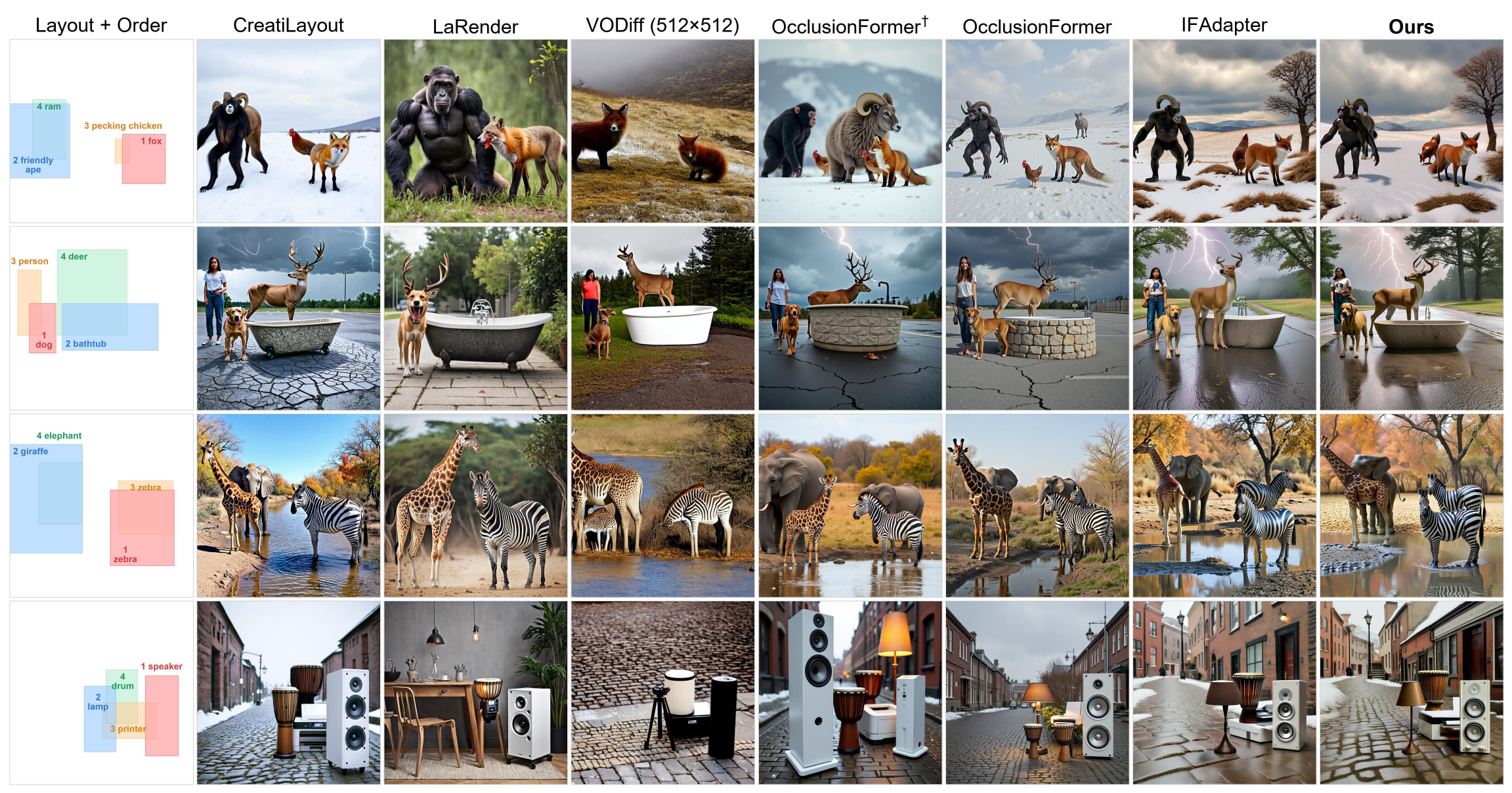}
\caption{
Qualitative comparison under overlapping layouts. $^\dagger$ denotes the official checkpoint without OccluLayout training.}
\label{fig:qualitative}
\end{figure*}


\subsection{Quantitative Comparison}
\label{subsec}

The quantitative results on OccluLayout-Bench are reported using three different MLLM evaluators. Although their absolute scores differ, they produce consistent method-level comparisons, as shown in Table~\ref{tab:main_results}. Specifically, OccluRank achieves the best Presence, Box mIoU, Color, Strict Pair, and Strict Image scores under all three evaluators. Its advantage is most pronounced on the strict occlusion metrics. Compared with IFAdapter, which shares the same SDXL backbone and per-instance feature construction, OccluRank improves Strict Pair by $7.38\%$-$10.61\%$ and Strict Image by $7.77\%$-$11.40\%$, while also consistently improving Presence and Box mIoU. This controlled comparison indicates that independently grounded instance features already provide effective localization, whereas rank-conditioned interaction before aggregation better resolves competing conditions in overlapping regions. 

In terms of attribute consistency, OccluRank achieves the best Color scores under all three evaluators and ranks first or second on Texture. For image quality, the reproduced OcclusionFormer achieves the best FID of $57.124$, but performs substantially worse on the other metrics. In contrast, OccluRank maintains a competitive FID of $62.746$ while achieving the strongest results on most metrics, providing the best overall balance between image quality and controllable occlusion-aware generation.

\subsection{Qualitative Analysis}
\label{subsec:qualitative}

Figure~\ref{fig:qualitative} compares different methods under identical inputs. Existing methods frequently omit requested instances, merge neighboring objects, deviate from the specified layouts, or produce incorrect visibility relationships in overlapping regions. In particular, LaRender, VODiff and OcclusionFormer often fail to preserve all instances, while CreatiLayout and IFAdapter generate more plausible compositions but still exhibit instance confusion or ordering errors under dense overlap. Our OccluRank consistently preserves the objects, aligns them with their designated regions, and realizes coherent visibility boundaries following the prescribed order across diverse animal and object scenes. These results demonstrate that rank-conditioned interaction before aggregation more effectively resolves competing instance conditions.

\begin{figure}[t]
    \centering
    \includegraphics[width=0.95\columnwidth]{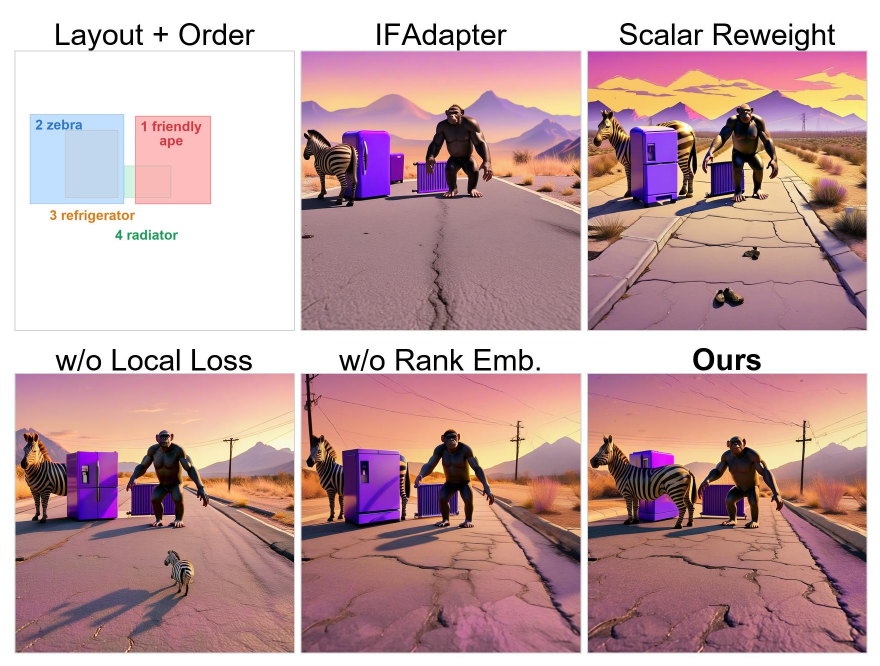}
    \caption{Qualitative comparison of the ablation variants under the same input layout and front-to-back order.}
    \label{fig:ablation}
\end{figure}


\subsection{Ablation Study}
\label{subsec:ablation}

We ablate the interaction strategy, the rank embedding, and the local objective. Figure~\ref{fig:ablation} qualitatively shows the changes in the image generated. It can be found that although the baseline and ablated models generate most requested categories, they exhibit ambiguous or reversed layering in the overlapping zebra-refrigerator and ape-radiator pairs. Some variants also introduce spurious content or unclear visibility boundaries. In particular, removing rank embeddings makes the interaction insensitive to the prescribed front and back roles, while removing the local objective weakens object structure within the conditioned regions. In contrast, the full model preserves all four instances and clearly places the zebra in front of the refrigerator and the ape in front of the radiator, demonstrating the complementary effects of order-aware interaction, rank conditioning, and local supervision.

\section{Conclusion}
We have presented OccluRank, a controllable occlusion-aware layout-to-image framework that adds an ordinal rank to each instance. Its OII module allows overlapping, rank-conditioned instance features to interact before aggregation, enabling explicit visibility control without additional geometric inputs or specialized inference. We also introduced OccluLayout, with geometry-derived occlusion and amodal annotations, and OccluLayout-Bench for synthetic controlled evaluation. Experiments with multiple MLLM evaluators show that OccluRank achieves the strongest results on most structured metrics while maintaining competitive attribute consistency and image quality.

\bibliography{aaai2027}


\clearpage
\appendix
\section{Additional Details on OccluLayout}

\subsection{Comparison with Existing Datasets}
\begin{table*}[t]
\centering
{
\small
\setlength{\tabcolsep}{1mm}

\begin{tabular}{@{}lcllccccccc@{}}
\toprule
Dataset
& Source
& \multicolumn{1}{c}{\#Img.}
& \multicolumn{1}{c}{\#Inst.}
& Res.
& Vocab.
& Inst. Text
& BBox
& Mask
& Z-order
& Amodal \\
\midrule

COCO 2017~\cite{lin2014microsoft}
& COCO
& $\approx 0.1\mathrm{M}$
& $\approx 0.88\mathrm{M}$
& Low
& 80
& Class
& $\checkmark$
& $\checkmark$
& --
& -- \\

InstaOrder~\cite{lee2022instance}
& COCO
& $\approx 0.1\mathrm{M}$
& $\approx 0.50\mathrm{M}$
& Low
& 80
& Class
& $\checkmark$
& $\checkmark$
& Human
& -- \\

COCOA~\cite{zhu2017semantic}
& COCO/BSDS
& $\approx 5.5\mathrm{K}$
& $\approx 69.0\mathrm{K}$
& Low
& 80
& Class
& $\checkmark$
& $\checkmark$
& Human
& Human \\

EliGen-Data~\cite{zhang2025eligen}
& EliGen
& $\approx 0.5\mathrm{M}$
& $\approx 1.26\mathrm{M}$
& High
& Open
& Phrase
& $\checkmark$
& $\times$
& --
& -- \\

LayoutSAM~\cite{zhang2025creatilayout}
& SA-1B
& $\approx 2.0\mathrm{M}$
& $\approx 10.7\mathrm{M}$
& High
& Open
& Phrase
& $\checkmark$
& $\times$
& --
& -- \\

SACap-1M~\cite{li2025seg2any}
& SA-1B
& $\approx 1.0\mathrm{M}$
& $\approx 5.88\mathrm{M}$
& High
& Open
& Phrase
& $\checkmark$
& $\checkmark$
& --
& -- \\

SA-Z~\cite{li2026occlusionformer}
& SA-1B
& $\approx 1.0\mathrm{M}$
& $\approx 5.69\mathrm{M}$
& High
& Open
& Phrase
& $\checkmark$
& $\checkmark$
& Predicted
& Predicted \\

SeeThrough3D~\cite{agrawal2026seethrough3d}
& Blender
& $\approx 51.8\mathrm{K}^{\dagger}$
& $\approx 0.17\mathrm{M}$
& High
& 48
& Class
& $\checkmark$
& $\times$
& Geometry
& -- \\

\midrule

\textbf{OccluLayout (Ours)}
& Blender
& $\approx 34.5\mathrm{K}$
& $\approx 0.12\mathrm{M}$
& High
& 71
& Phrase
& $\checkmark$
& $\checkmark$
& Geometry
& Geometry \\

\bottomrule
\end{tabular}
}

\caption{
Statistical comparison of layout and occlusion-related datasets.
Resolution is classified as High when the image's long edge exceeds
1000 pixels.
BBox and Mask indicate whether the released dataset provides
bounding-box and object-level mask annotations, respectively.
Z-order and Amodal report how the corresponding annotations are
obtained: ``Human'' denotes manual annotation, ``Predicted'' denotes
model prediction, ``Geometry'' denotes direct derivation from the
underlying 3D scene geometry, and ``--'' indicates that the annotation
is unavailable.
Vocabulary denotes the number of distinct instance labels; duplicate
asset-name suffixes are merged when counting OccluLayout.
$^{\dagger}$ The released SeeThrough3D data include raw Blender renders
and depth-to-image variants, and multiple images may share the same
underlying 3D layout.
}
\label{tab:dataset_comparison}
\end{table*}

We comprehensively compare our OccluLayout dataset with representative layout and occlusion image generation datasets in Table~\ref{tab:dataset_comparison}. COCO-based datasets provide bounding boxes and masks but are limited to low-resolution images and 80 class-level categories. COCO 2017 contains no occlusion annotations, while InstaOrder and COCOA rely on manual annotation for Z-order or amodal geometry. Recent layout datasets, including EliGen-Data, LayoutSAM, and SACap-1M, offer high-resolution images and open-vocabulary phrase descriptions, but do not provide explicit Z-order or amodal annotations. SA-Z includes both annotations, but obtains them through auxiliary model prediction from partially occluded images, which may introduce errors in the inferred order and hidden object regions. SeeThrough3D derives Z-order from controllable 3D geometry, but provides only 48 class-level categories and lacks object-level masks and amodal annotations.

Although our OccluLayout is smaller than large-scale SA-1B-derived datasets, it is designed to provide complete and reliable supervision for occlusion-aware layout-to-image generation. Among the compared datasets, it is the only one that jointly provides high-resolution images, phrase-level instance descriptions, bounding boxes, object masks, and geometry-derived Z-order and amodal annotations. Deriving the order, boxes, and masks from the same underlying 3D scene avoids dependence on manual labeling or auxiliary prediction and maintains their geometric consistency. The 71-category vocabulary and fine-grained instance descriptions further support attribute-controllable generation beyond category-level supervision.

\subsection{Detailed Dataset Construction}

\paragraph{Controllable scene generation.}
We construct OccluLayout in Blender~\cite{blender2018} using public 3D assets organized into 71 instance categories, covering furniture, appliances, vehicles, musical instruments, animals, and other everyday objects. For each scene, we randomly select two to five objects and place them on a shared ground plane to avoid floating objects and implausible support relationships. Their positions, scales, orientations, and camera viewpoints are varied to create different spatial arrangements and image-plane overlaps. The camera is automatically adjusted to keep all objects within the field of view. We additionally test intersections between the 3D axis-aligned bounding boxes of different objects and resample scenes containing physical interpenetration.

\paragraph{Amodal annotation and order extraction.}
For each accepted scene, we render a $1024\times1024$ RGB image and an amodal mask for every instance. To obtain the amodal mask, we hide all other objects, assign the target object a white emissive material, and render it against a black background. The tight rectangle enclosing the complete projected mask is used as its 2D amodal bounding box. Unlike a visible bounding box, this annotation describes the complete intended extent of an instance, including its occluded region.

The front-to-back order is derived from the relative object positions in the camera coordinate system. Instances are stored according to this order, with a smaller index indicating a more foreground position. Consequently, the RGB image, amodal masks, bounding boxes, and order annotations originate from the same scene geometry rather than being inferred retrospectively from the rendered image. The underlying depth, camera parameters, and 3D geometry are used only during dataset construction and are not provided to the generation model.

\paragraph{Automatic quality filtering.}
We apply two filters to retain recognizable instances under meaningful occlusion. First, a scene is discarded if the bounding-box area of any instance is below $1\%$ of the image area. Second, we measure how much of each background instance is covered by instances placed in front of it. Let $\mathcal{A}_i$ denote the pixel set of the amodal mask of instance $i$. Its occlusion ratio is defined as
\begin{equation}
\rho_i=
\frac{
\left|
\mathcal{A}_i
\cap
\left(
\bigcup_{j:r_j<r_i}\mathcal{A}_j
\right)
\right|
}{
|\mathcal{A}_i|
}.
\label{eq:dataset_occlusion_ratio}
\end{equation}
We reject a scene when $\max_i\rho_i>0.70$. These filters exclude extremely small or almost entirely invisible instances while retaining explicit foreground-background interactions.

\paragraph{Background and appearance diversification.}
Raw 3D renders have limited background and appearance diversity. We therefore use FLUX.2 [Klein] 4B to edit scene backgrounds and selected object appearances. The background prompt pool contains 13,728 combinations constructed from 143 scene types, four seasons, four weather conditions, and six times of day. One combination is randomly selected for each rendered image. The editing instruction requires the positions, scales, and shapes of all foreground objects to remain unchanged. For eligible object categories, we additionally perform randomized color or texture editing. Colors are sampled from 15 candidates and textures from 10 candidates, while some objects retain their original appearances. These edits are constrained to preserve the object geometry, spatial placement, and front-to-back relationship.

\paragraph{Image validation and text annotation.}
After editing, we use Qwen3-VL-235B-A22B-Instruct~\cite{bai2025qwen3vl} to verify that every instance remains aligned with its annotated bounding box. Images containing object drift, disappearance, or category changes are discarded. This validation reduces inconsistencies introduced by image editing. Qwen3-VL-235B-A22B-Instruct also generates a global caption and fine-grained descriptions for individual instances. For each instance, the model receives the complete image, its category name, and its bounding box. It describes only intrinsic properties, including color, material, texture, shape, pose, and action. Inter-object relations, such as contact, support, containment, and occlusion, are excluded to prevent the description from implicitly revealing the target order. The global caption summarizes the principal instances, their two-dimensional arrangement, and the background environment. 

Finally, each retained sample contains one RGB image, one global caption, instance-level descriptions, 2D amodal bounding boxes, a front-to-back ordered instance list, and auxiliary amodal masks. The current model does not use the amodal masks as input conditions or training supervision.

\subsection{Dataset Statistics}
\label{app:dataset_statistics}

The complete OccluLayout collection contains 34,496 images and 117,433 annotated instances. We randomly select 1,000 images containing 3,386 instances to form OccluLayout-Bench and use the remaining 33,496 images with 114,047 instances for training. Both splits contain scenes with two to five objects. Detailed statistics are reported in Table~\ref{tab:occlulayout_statistics}.

\begin{table}[t]
\centering
\begin{tabular}{@{}lcc@{}}
\toprule
Statistic & Train & Test \\
\midrule
\# Images
& 33,496
& 1,000 \\

\# Instances
& 114,047
& 3,386 \\

Avg. Instances / Image
& 3.40
& 3.39 \\

Images with Box Overlap
& 78.8\%
& 77.2\% \\

Mean Max Pairwise Box IoU
& 0.2215
& 0.2190 \\

Mean Relative BBox Area
& 0.0630
& 0.0631 \\

\bottomrule
\end{tabular}
\caption{
Statistics of the OccluLayout training and test splits.
Box overlap denotes images containing at least one pair of
overlapping bounding boxes.
}
\label{tab:occlulayout_statistics}
\end{table}

The training split contains 7,307, 11,066, 9,380, and 5,743 images with two, three, four, and five instances, respectively. The corresponding numbers in OccluLayout-Bench are 219, 339, 279, and 163. Their average numbers of instances per image are 3.40 and 3.39, indicating closely matched instance-count distributions. To characterize layout overlap, we compute the maximum pairwise bounding-box IoU in each image. Overlapping boxes occur in $78.8\%$ of the training images and $77.2\%$ of the benchmark images, while their mean maximum pairwise IoUs are 0.2215 and 0.2190. Their mean relative bounding-box areas are also similar, at 0.0630 and 0.0631. These statistics indicate that the two splits have comparable object-scale and overlap distributions.

OccluLayout-Bench contains 772 images with at least one overlapping bounding-box pair, yielding 1,762 pairs for occlusion evaluation. Among them, 503 images contain at least two overlapping pairs, accounting for $50.3\%$ of the benchmark. It therefore covers both isolated pairwise occlusion and multi-instance scenes containing several simultaneous visibility relationships.

\section{Additional Details on OccluLayout-Bench}
OccluLayout-Bench is designed to evaluate whether a model simultaneously preserves the requested instances, follows their layouts and attributes, and realizes the specified occlusion order. It therefore assesses five complementary dimensions: instance presence (Presence), spatial adherence (Box mIoU), attribute consistency (Color and Texture), occlusion-order adherence (Strict Pair and Strict Image), and overall image quality (FID). Presence, Box mIoU, Color and Texture metrics are computed over the 3,386 target instances, while the order metrics Strict Pair \& Strict Image cover 1,762 overlapping pairs from 772 images. FID is computed over all 1,000 benchmark images.

We employ Qwen3-VL-235B-A22B-Instruct, Qwen3-VL-32B-Instruct,
and GLM-4.6V as the three main MLLM evaluators. The two
Qwen3-VL models use their Instruct checkpoints, while GLM-4.6V
is evaluated with reasoning mode set to \texttt{none}. We
additionally report supplementary ablation results using
Qwen3.7-Plus with reasoning mode set to \texttt{none}. All
evaluators receive identical image inputs, crop regions, task
prompts, and structured output formats. Their results are reported
separately rather than averaged, allowing us to examine whether
the method-level conclusions remain consistent across evaluator
choices. All MLLM-based metrics are reported as percentages.

Next, we introduce the definition of each metric.
\begin{itemize}
    \item \textbf{Presence:} For each target instance, we crop its prescribed bounding-box region and ask the evaluator whether an object of the requested category is recognizable. A partially occluded object is counted as present if its category remains identifiable, whereas a missing or unrecognizable instance is counted as a failure. We compute
   \begin{equation}
  \mathrm{Presence}= \frac{N_{\mathrm{present}}}{N_{\mathrm{instance}}}\times 100,
   \end{equation}
where $N_{\mathrm{present}}$ denotes the number of recognizable target instances. This metric evaluates whether the requested instances are successfully generated within their conditioned regions, independently of precise box alignment.

\item \textbf{Box mIoU:} To evaluate the realized object locations without revealing the target coordinates, the evaluator receives the full generated image and the requested category names and predicts candidate bounding boxes. Within each category, we perform one-to-one optimal matching between the predicted and ground-truth boxes, ensuring that a single prediction cannot be assigned to multiple target instances. Unmatched ground-truth instances receive an IoU of zero. We compute
\begin{equation}
\mathrm{Box\ mIoU}=\frac{1}{N_{\mathrm{instance}}}\sum_{i=1}^{N_{\mathrm{instance}}}\operatorname{IoU}\left(B_i^{\mathrm{gt}},B_i^{\mathrm{pred}}\right)\times 100.
\end{equation}
Because the input boxes represent amodal object extents, whereas the predicted boxes are estimated from visible image evidence, Box mIoU provides a conservative measure of spatial adherence under occlusion.

\item \textbf{Color and Texture:} For each target instance whose description explicitly specifies the evaluated attribute (color or texture), the evaluator assigns a score of 2 for a clear match, 1 for a partial match, and 0 for a mismatch. A missing or unrecognizable instance also receives zero. For each attribute, instances without a corresponding annotation are excluded from the evaluation. We compute
\begin{equation}
\mathrm{Attribute}=\frac{\sum_{i=1}^{N_{\mathrm{specified}}} s_i}{2N_{\mathrm{specified}}}\times 100,
\end{equation}
where $s_i\in\{0,1,2\}$ is the score for instance $i$, and $N_{\mathrm{specified}}$ is the number of instances for which the evaluated attribute is explicitly specified. We report \textbf{Color} and \textbf{Texture} separately, with Texture covering both texture and material descriptions. By assigning zero to absent instances, this strict protocol jointly evaluates successful instance generation and preservation of the requested appearance.

\item \textbf{Strict Pair and Strict Image:} For each pair of overlapping input boxes, we crop their union region and evaluate it in two stages. The evaluator first determines whether both requested instances are recognizable. Only when both are present does it predict their visible front-to-back relation, without access to the specified order. We report \textbf{Strict Pair}, the percentage of all overlapping pairs for which both instances are recognizable and the predicted relation matches the specified order. Missing or unrecognizable instances and incorrect relations are all counted as failures. We further report \textbf{Strict Image}, the percentage of images with overlapping boxes for which every overlapping pair satisfies the Strict Pair criterion. Strict Pair therefore jointly evaluates instance preservation and pairwise ordering, while Strict Image measures whether all prescribed occlusion relations in a multi-instance scene are simultaneously realized.

\end{itemize}

\paragraph{Overall image quality.} 
We compute Fr'echet Inception Distance (FID)~\cite{heusel2017gans} between the 1,000 generated images and their corresponding OccluLayout-Bench reference images. FID is independent of the MLLM evaluators and provides a complementary distribution-level measure of overall image quality.

Together, these metrics distinguish different failure sources in occlusion-aware generation. Presence and Box mIoU measure whether instances are generated and correctly placed, Color and Texture assess whether their specified attributes are preserved, and the strict order metrics jointly penalize instance omission and incorrect visibility relationships. This protocol prevents a model from receiving a favorable occlusion score by generating only an easily ordered subset of the requested objects.

\subsection{Complementary Occlusion and Depth Evaluation}
\label{subsec:complementary_order_evaluation}

To complement the MLLM-based evaluation, we implement a specialized
order-prediction protocol following the occlusion- and depth-order
evaluation used in OcclusionFormer~\cite{li2026occlusionformer}.
This protocol assesses the realized pairwise relationships using
dedicated order-prediction networks rather than an MLLM.

For every pair of overlapping input bounding boxes, the prescribed
relation is determined directly from the front-to-back annotation
order. Since instances in OccluLayout are stored from foreground to
background, for a pair of instances $i<j$, instance $i$ is expected to
appear in front of instance $j$. We first use SAM 3~\cite{carion2026sam3}, conditioned on the
instance category and its input bounding box, to obtain a visible
instance mask from the generated image. The RGB image and the two
instance masks are subsequently passed to InstaOrderNet and
InstaDepthNet~\cite{lee2022instance} to predict the realized occlusion
and depth relations, respectively. If either requested instance cannot
be segmented, the corresponding pair is counted as incorrect.

For occlusion order, we report the standard F1 score
\begin{equation}
\mathrm{Occ.}
=
\frac{2PR}{P+R},
\qquad
P=\frac{\mathrm{TP}}{\mathrm{TP}+\mathrm{FP}},
\qquad
R=\frac{\mathrm{TP}}{\mathrm{TP}+\mathrm{FN}},
\end{equation}
where the statistics are accumulated over the predicted and prescribed
pairwise occlusion directions of all overlapping-box pairs. Higher
values indicate better adherence to the requested occlusion order.

For depth order, we report the weighted human disagreement rate
(WHDR),
\begin{equation}
\mathrm{Dep.}
=
\frac{
\sum_{(i,j)\in\mathcal{P}}
w_{ij}\,
\mathbb{I}\!\left[\hat{d}_{ij}\neq d_{ij}\right]
}{
\sum_{(i,j)\in\mathcal{P}} w_{ij}
},
\end{equation}
where $\hat{d}_{ij}$ and $d_{ij}$ denote the predicted and prescribed
depth relations, respectively. Since OccluLayout-Bench does not contain
human confidence weights, we assign a uniform weight $w_{ij}=1$ to
every evaluated pair. Lower values indicate better depth-order
consistency.

\begin{table}[t]
\centering
\small
\setlength{\tabcolsep}{3.5mm}

\begin{tabular}{@{}l|c|cc@{}}
\toprule
Method
& Venue
& Occ. $\uparrow$
& Dep. $\downarrow$ \\
\midrule

CreatiLayout
& ICCV'2025
& 0.7552
& 0.2244 \\

LaRender
& ICCV'2025
& 0.5908
& 0.4392 \\

OcclusionFormer$^{\dagger}$
& ICML'2026
& \underline{0.8184}
& 0.2126 \\

OcclusionFormer
& ICML'2026
& 0.7531
& 0.2423 \\

VODiff
& CVPR'2025
& 0.7065
& 0.2548 \\

IFAdapter
& ICCV'2025
& 0.7987
& \underline{0.1993} \\

\textbf{Ours}
& --
& \textbf{0.8577}
& \textbf{0.1844} \\

\bottomrule
\end{tabular}

\caption{
Additional quantitative comparison on OccluLayout-Bench using
occlusion-order F1 (Occ.) and depth-order WHDR (Dep.).
VODiff is evaluated at $512{\times}512$ and all other methods at
$1024{\times}1024$.
$^{\dagger}$ denotes the official checkpoint without further training on
OccluLayout.
}
\label{tab:occ_dep_results}
\end{table}

As shown in Table~\ref{tab:occ_dep_results}, OccluRank achieves the
best performance on both complementary metrics. It obtains an
occlusion-order F1 of $0.8577$ and a depth-order WHDR of $0.1844$,
consistently outperforming the compared layout-to-image and
occlusion-aware generation methods. These results complement the MLLM
evaluation and provide additional evidence that OccluRank realizes the
specified foreground-to-background relationships.

\section{Baselines and Implementation Details}
\label{subsec:implementation_details}

\subsection{Baselines}
We compare OccluRank with representative layout-to-image and occlusion-aware generation methods, including CreatiLayout~\cite{zhang2025creatilayout}, IFAdapter~\cite{wu2025ifadapter}, LaRender~\cite{zhan2025larender}, VODiff~\cite{liang2025vodiff}, and OcclusionFormer~\cite{li2026occlusionformer}. CreatiLayout provides fine-grained instance-level layout control, while IFAdapter is our direct architectural baseline and aggregates independently constructed instance features through gated semantic fusion. LaRender performs rendering-inspired latent composition, VODiff controls visibility through order-dependent denoising, and OcclusionFormer learns Z-order-aware volumetric feature composition.

We use the released backbone configuration of each method. CreatiLayout is built on Stable Diffusion 3~\cite{esser2024scaling}; IFAdapter and OccluRank use SDXL; LaRender uses the SDXL-based IterComp checkpoint~\cite{zhang2025itercomp}; VODiff uses the official GLIGEN text-box checkpoint~\cite{li2023gligen} and generates images at $512\times512$; and OcclusionFormer uses FLUX.1-dev. OccluRank and IFAdapter share the same backbone and sampling configuration, providing a controlled comparison of the proposed order-aware interaction. For the other methods, we follow their official or recommended inference settings, including denoising steps, guidance scales, schedulers, and input resolutions.

All methods that require training or fine-tuning use the same OccluLayout training split. Because the official training code for OcclusionFormer is unavailable, we implement a best-effort reproduction following its published specification. To account for a potential reproduction gap, we additionally evaluate the authors' released checkpoint without further training on OccluLayout, denoted as \textit{OcclusionFormer$^\dagger$}.

\subsection{Implementation Details}
OccluRank is initialized from pretrained SDXL and IFAdapter checkpoints. The diffusion backbone remains frozen, while the per-instance feature construction, ordinal rank embeddings, OII block, and residual injection parameters are optimized. For each instance, we extract three hidden-state layers from the SDXL text encoders and obtain four Resampler query tokens from each layer. After flattening the layer and query dimensions and concatenating one EOT token, each instance is represented by $L_T=13$ layout tokens. We cap each layout at $O_{\max}=5$ instances and use learned ordinal-rank embeddings. The OII module contains one Transformer layer operating along the instance dimension.

The OII-enhanced instance signal is injected into the cross-attention layers of Transformer sub-blocks 0--3 in the SDXL middle block and the lowest-resolution upsampling block (\texttt{up\_blocks.0}); all other attention layers remain unchanged. The adapter scale is fixed to $s=1.0$, while each learnable residual gate $\alpha$ is initialized to zero. Training is performed at $1024\times1024$ resolution on a single NVIDIA RTX PRO 6000 GPU. We use AdamW with a learning rate of $1\times10^{-4}$, an effective batch size of 160, and 1,500 optimization steps under bf16 mixed precision. We independently apply classifier-free condition dropout to the global prompt and the complete set of local instance conditions with probabilities of 0.30 and 0.15, respectively. Dropping the local conditions removes both the instance-text embeddings and their associated spatial attention masks. The balance weight is set to $\lambda=2.0$, and the numerical-stability constant is set to $\delta=10^{-8}$.

During inference, OccluRank generates images at $1024\times1024$ resolution using 30 denoising steps, a classifier-free guidance scale of 7.5, and a fixed random seed of 42. We generate one image for each layout in OccluLayout-Bench.

\begin{table}[t]
\centering
\small
\setlength{\tabcolsep}{3.5mm}

\begin{tabular}{@{}l|c|cc@{}}
\toprule
Variant
& Venue
& Occ. $\uparrow$
& Dep. $\downarrow$ \\
\midrule

IFAdapter
& ICCV'2025
& 0.7987
& 0.1993 \\

Scalar Reweight
& --
& 0.8127
& 0.2073 \\

w/o Local Loss
& --
& \underline{0.8400}
& 0.1866 \\

w/o Rank Emb.
& --
& 0.8279
& \underline{0.1859} \\

\textbf{Ours}
& --
& \textbf{0.8577}
& \textbf{0.1844} \\

\bottomrule
\end{tabular}

\caption{
Ablation study on OccluLayout-Bench using occlusion-order F1
(Occ.) and depth-order WHDR (Dep.).
}
\label{tab:occ_dep_ablation}
\end{table}
\begin{table*}[t]
\centering
\small
\setlength{\tabcolsep}{1mm}

\begin{tabular}{@{}c|l|c|ccccccc@{}}
  \toprule
  Evaluator (MLLM)
  & \multicolumn{1}{c|}{Variant}
  & \multicolumn{1}{c|}{Venue}
  & Presence $\uparrow$
  & Box mIoU $\uparrow$
  & Color $\uparrow$
  & Texture $\uparrow$
  & Strict Pair $\uparrow$
  & Strict Image $\uparrow$
  & FID $\downarrow$ \\
  \midrule

  \multirow[c]{5}{*}{\shortstack[c]{Qwen3-VL-235B-\\A22B}}
  & IFAdapter
  & ICCV'2025
  & 90.17
  & 60.80
  & 83.46
  & 74.84
  & 66.06
  & 51.55
  & 62.923 \\

  & Scalar Reweight
  & --
  & 90.37
  & 59.74
  & \underline{84.21}
  & \underline{75.78}
  & 71.40
  & 57.38
  & \textbf{62.277} \\

  & w/o Local Loss
  & --
  & \underline{90.96}
  & 63.10
  & 84.09
  & 75.39
  & 72.81
  & 56.61
  & \underline{62.322} \\

  & w/o Rank Emb.
  & --
  & 90.79
  & \textbf{63.32}
  & 83.93
  & 74.18
  & \underline{73.89}
  & \underline{59.33}
  & 64.332 \\

  & \textbf{Ours}
  & --
  & \textbf{91.58}
  & \underline{63.20}
  & \textbf{85.23}
  & \textbf{76.02}
  & \textbf{76.45}
  & \textbf{62.44}
  & 62.746 \\

  \midrule

  \multirow[c]{5}{*}{Qwen3-VL-32B}
  & IFAdapter
  & ICCV'2025
  & 91.44
  & 60.87
  & 82.59
  & \textbf{73.19}
  & 66.46
  & 51.42
  & 62.923 \\

  & Scalar Reweight
  & --
  & 91.38
  & 60.72
  & \underline{83.59}
  & \underline{73.00}
  & 71.28
  & 55.96
  & \textbf{62.277} \\

  & w/o Local Loss
  & --
  & 91.88
  & \textbf{64.36}
  & 83.28
  & 71.98
  & 71.51
  & 54.79
  & \underline{62.322} \\

  & w/o Rank Emb.
  & --
  & \underline{92.06}
  & 63.95
  & 83.06
  & 72.62
  & \underline{73.21}
  & \underline{56.87}
  & 64.332 \\

  & \textbf{Ours}
  & --
  & \textbf{92.50}
  & \underline{64.23}
  & \textbf{83.92}
  & 72.92
  & \textbf{77.07}
  & \textbf{62.82}
  & 62.746 \\

  \midrule

  \multirow[c]{5}{*}{GLM-4.6V}
  & IFAdapter
  & ICCV'2025
  & 91.79
  & 61.68
  & 86.64
  & 72.13
  & 59.93
  & 44.69
  & 62.923 \\

  & Scalar Reweight
  & --
  & 91.58
  & 60.93
  & 86.65
  & \underline{73.67}
  & 64.70
  & \underline{51.68}
  & \textbf{62.277} \\

  & w/o Local Loss
  & --
  & \underline{92.56}
  & \textbf{64.77}
  & \underline{87.53}
  & \textbf{73.76}
  & \underline{65.83}
  & 50.91
  & \underline{62.322} \\

  & w/o Rank Emb.
  & --
  & 92.32
  & 64.10
  & 87.08
  & 73.59
  & 64.98
  & 50.26
  & 64.332 \\

  & \textbf{Ours}
  & --
  & \textbf{92.85}
  & \underline{64.13}
  & \textbf{87.79}
  & 73.23
  & \textbf{67.31}
  & \textbf{52.46}
  & 62.746 \\

  \midrule

  \multirow[c]{5}{*}{Qwen3.7-Plus}
  & IFAdapter
  & ICCV'2025
  & 91.32
  & 62.38
  & 83.70
  & \textbf{80.74}
  & 57.21
  & 40.93
  & 62.923 \\

  & Scalar Reweight
  & --
  & 91.08
  & 61.19
  & 84.31
  & \underline{80.72}
  & 61.46
  & 44.56
  & \textbf{62.277} \\

  & w/o Local Loss
  & --
  & 91.73
  & \textbf{64.86}
  & 84.30
  & 80.10
  & \underline{65.32}
  & 46.50
  & \underline{62.322} \\

  & w/o Rank Emb.
  & --
  & \underline{92.17}
  & 64.36
  & \underline{84.53}
  & 79.66
  & 65.15
  & \underline{47.41}
  & 64.332 \\

  & \textbf{Ours}
  & --
  & \textbf{92.65}
  & \underline{64.60}
  & \textbf{85.15}
  & 80.42
  & \textbf{68.84}
  & \textbf{51.17}
  & 62.746 \\

  \bottomrule
\end{tabular}

\caption{
Ablation study on OccluLayout-Bench using four MLLM evaluators.
}
\label{tab:ablation}
\end{table*}

\section{Additional Ablation Analysis}
\label{subsec:additional_ablation}

We provide additional quantitative ablations to complement the qualitative analysis in the main paper. All variants share the same initialization, training configuration, and inference settings. \textit{Scalar Reweight} replaces OII with order-conditioned scalar weights for instance aggregation, while \textit{w/o Local Loss} and \textit{w/o Rank Emb.} remove the local denoising objective and ordinal rank embeddings, respectively.

The complementary order metrics exhibit a consistent trend with the
MLLM-based ablations. Scalar Reweight improves Occ. over IFAdapter,
showing that incorporating the prescribed order during aggregation is
beneficial, but it remains below the full interaction-based model.
Removing either the local objective or the ordinal rank embedding also
degrades occlusion-order accuracy. The full model achieves the highest
Occ. score of $0.8577$ and the lowest Dep. value of $0.1844$, supporting
the joint contribution of explicit rank conditioning, order-aware
instance interaction, and localized denoising supervision.

Scalar Reweight improves Strict Pair and Strict Image over
IFAdapter by $4.25$--$5.34$ and $3.63$--$6.99$ percentage
points across the four evaluators, respectively, confirming the
benefit of introducing order information during aggregation.
Nevertheless, the full model further improves these metrics by
$2.61$--$7.38$ and $0.78$--$6.86$ points. This gap supports
our interaction-before-aggregation design: scalar weights only
regulate the contributions of independently constructed features,
whereas OII updates their high-dimensional content according to
the competing instances and prescribed order.

Removing the local objective (i.e., $\lambda=0$) has only a minor
effect on Box mIoU and produces marginal improvements under three
evaluators. However, it reduces Strict Pair by $1.48$--$5.56$
points and Strict Image by $1.55$--$8.03$ points. The local
objective therefore contributes primarily to recognizable instance
structures and coherent boundaries within conditioned regions,
rather than coarse object placement. It complements OII by
encouraging the learned interaction to be expressed as visually
distinguishable occlusion relationships.

Removing the rank embedding leaves instance presence and spatial
alignment largely unchanged, but decreases Strict Pair and Strict
Image by $2.33$--$3.86$ and $2.20$--$5.95$ points, respectively.
Although the instance-dimension Transformer can still exchange
semantic information, it lacks an explicit asymmetric signal for
distinguishing the prescribed front and back roles.

\begin{figure}[t]
    \centering
    \includegraphics[width=\columnwidth]{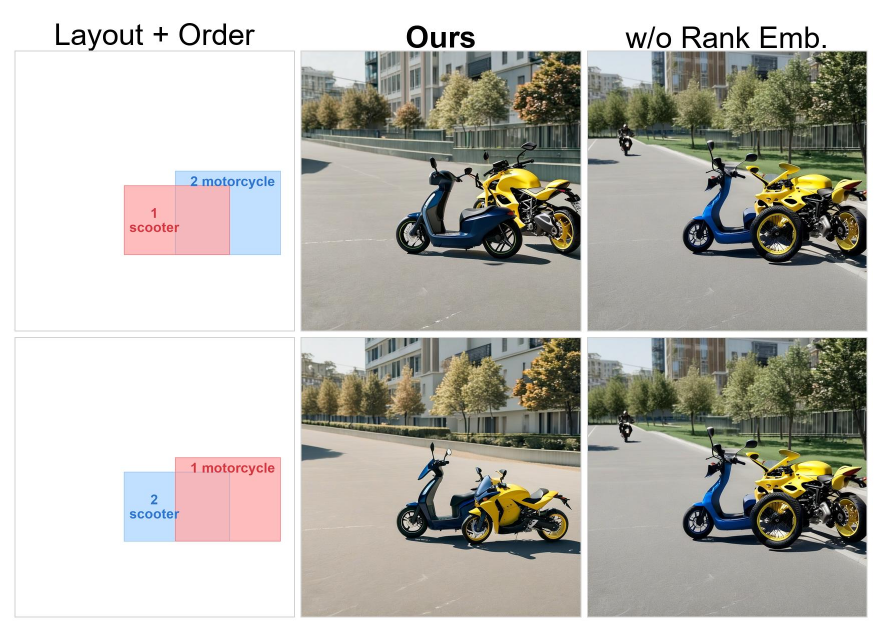}
    \caption{Qualitative comparison under reversed instance order. The variant without rank embeddings fails to respond to the order change, while our full model successfully controls the front-to-back relationship.}
    \label{fig:reverse_ablation}
\end{figure}

\paragraph{Responsiveness to Reversed Occlusion Orders.}
To verify that OccluRank genuinely follows the specified order rather than relying on learned semantic or layout priors, we construct paired conditions that differ only in their front-to-back order while preserving the global prompt, instance descriptions, and bounding boxes. As shown in Fig.~\ref{fig:reverse_ablation}, OccluRank reverses the realized covering relationship in accordance with the input order, whereas \textit{w/o Rank Emb.} produces nearly unchanged occlusion patterns. This controlled comparison demonstrates that rank embeddings provide the asymmetric signal required for order-responsive generation, which cannot be reliably achieved through instance interaction alone.

\section{Generalization to Unseen Categories}
\label{sec:unseen_categories}

\begin{figure*}[!t]
\centering
\includegraphics[width=0.96\textwidth]{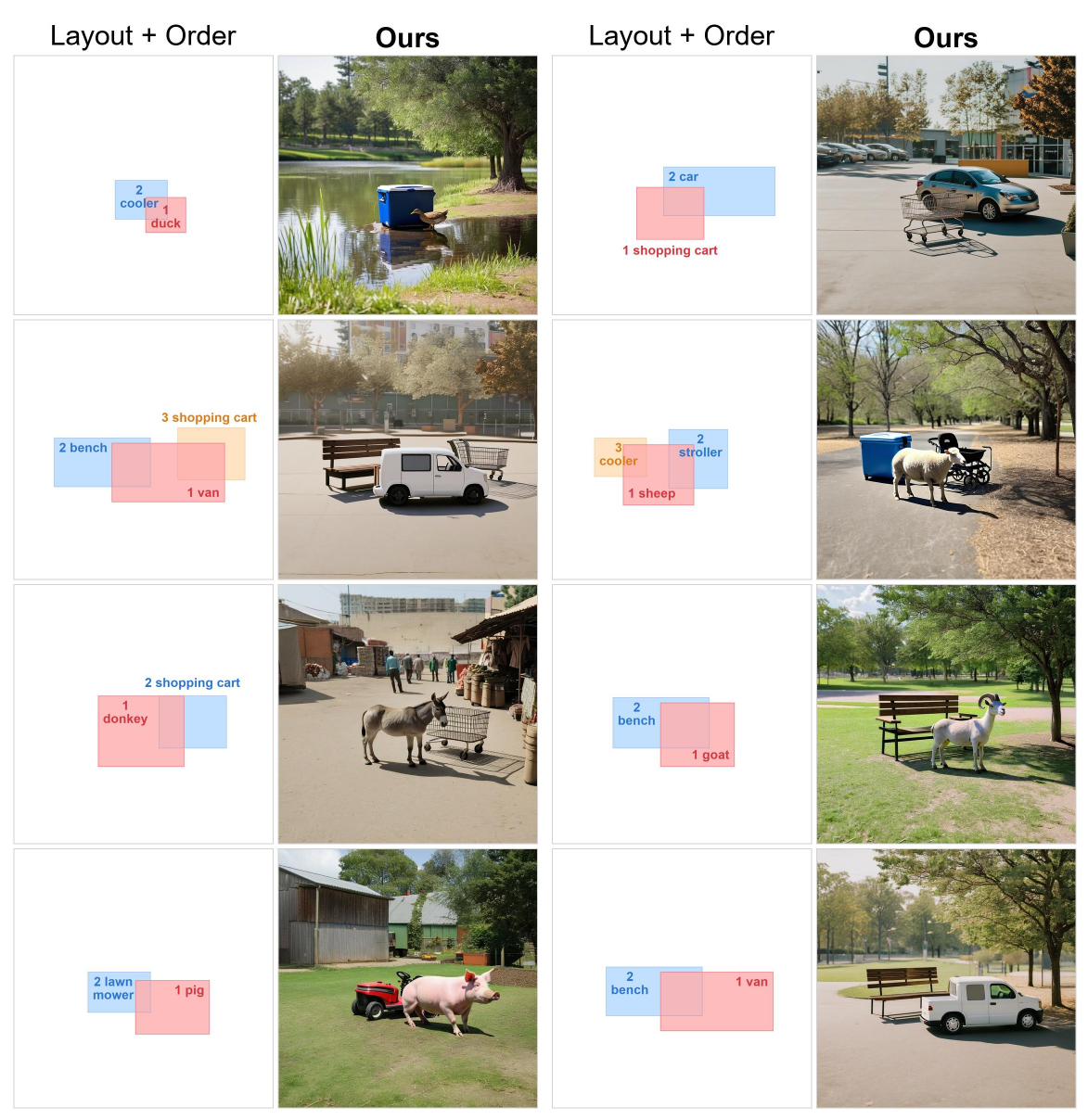}
\caption{
Qualitative examples on categories unseen during training. Numeric prefixes denote the prescribed front-to-back order. OccluRank generalizes its occlusion control to unseen categories, preserving the requested instances while correctly realizing the specified occlusion relationships.}
\label{fig:unseen}
\end{figure*}

We further examine whether the occlusion-control capability of
OccluRank generalizes beyond the 71 instance categories included
in OccluLayout. We construct additional overlapping layouts using
categories absent from the training set and assign explicit
front-to-back orders to their instances. As shown in
Fig.~\ref{fig:unseen}, OccluRank successfully preserves the
requested instances and realizes the specified occlusion
relationships. These results demonstrate that the learned
order-aware interaction is not restricted to the object categories
observed during training, but generalizes to occlusion relationships
between unseen categories.

\end{document}